\documentclass[10pt,twocolumn,letterpaper]{article}

\usepackage{cvpr}              
\usepackage{setspace}
\usepackage{algorithm}  
\usepackage{algpseudocode}
\usepackage{float}
\usepackage[T1]{fontenc}

\definecolor{cvprblue}{rgb}{0.21,0.49,0.74}
\usepackage[pagebackref,breaklinks,colorlinks,allcolors=cvprblue]{hyperref}

\title{Towards Efficient and Robust Moment Retrieval System: A Unified Framework for Multi-Granularity Models and Temporal Reranking}

\author{
\thanks{All authors contributed equally to this paper. \\ 
This research is fully supported by AI VIETNAM \cite{aivietnamVit}.}
Huu-Loc Tran $^{1*}$ \quad Tinh-Anh Nguyen-Nhu $^{2*}$ \quad  Huu-Phong Phan-Nguyen $^{1*}$ \\ \quad Tien-Huy Nguyen$^{1}$ \quad Nhat-Minh Nguyen-Dich$^{3}$ \quad Anh Dao $^{4}$  \\ \quad Huy-Duc Do $^{5}$ \quad Quan Nguyen $^{6}$\quad Hoang M. Le $^{7}$\quad Quang-Vinh Dinh $^{8}$\\ \\
$^{1}$ University of Information Technology, VNU-HCM, Vietnam  \\
$^{2}$ Ho Chi Minh University of Technology, VNU-HCM, Vietnam \\
$^{3}$ Hanoi University of Science and Technology, Hanoi, Vietnam \\
$^{4}$ Michigan State University, USA \\
$^{5}$ National Economics University, Hanoi, Vietnam \\
$^{6}$ Posts and Telecommunications Institute of Technology, Hanoi, Vietnam \\
$^{7}$ York University, Canada\\
$^{8}$ AI VIETNAM Lab \\
{\tt\small 22520567@gm.uit.edu.vn}}

\begin{document}
\maketitle
\begin{abstract}
Long-form video understanding presents significant challenges for interactive retrieval systems, as conventional methods struggle to process extensive video content efficiently. Existing approaches often rely on single models, inefficient storage, unstable temporal search, and context-agnostic reranking, limiting their effectiveness. This paper presents a novel framework to enhance interactive video retrieval through four key innovations: (1) an ensemble search strategy that integrates coarse-grained (CLIP) and fine-grained (BEIT3) models to improve retrieval accuracy, (2) a storage optimization technique that reduces redundancy by selecting representative keyframes via TransNetV2 and deduplication, (3) a temporal search mechanism that localizes video segments using dual queries for start and end points, and (4) a temporal reranking approach that leverages neighboring frame context to stabilize rankings. Evaluated on known-item search and question-answering tasks, our framework demonstrates substantial improvements in retrieval precision, efficiency, and user interpretability, offering a robust solution for real-world interactive video retrieval applications. 
\end{abstract} 
\vspace{-1.5em}
\section{Introduction}
\label{sec:intro}

Recent advances in deep learning have significantly improved core vision tasks such as recognition, domain adaptation, and visual question answering \cite{nguyensemi, nguyen2025enhancing, ngo2024dual, 10661057, nguyen2024emotic}, paving the way for more capable video understanding systems. However, most existing approaches \cite{karpathy2014large, tran2015learning, feichtenhofer2019slowfast, wang2018non} are designed for short video clips—typically only seconds to a few minutes long—making them ill-suited for real-world scenarios involving long-form videos \cite{wu2021longformvideounderstanding}. This gap presents a major challenge for applications like content-based video retrieval and surveillance, where processing hours-long content is essential.

Interactive video retrieval \cite{ma2022interactive, liang2023simple, 10.1145/3628797.3629011} has emerged as a promising solution by combining automated analysis with human input. 
Recent advancements, fueled by deep learning and validated in competitions like TRECVID \cite{TRECVID} and VBS \cite{VBS}, demonstrate that human-computer collaboration not only refines automated results in real time but also substantially enhances search effectiveness.

The core challenge lies in translating a user's information need into an efficient search process that identifies relevant segments within massive video archives. Current state-of-the-art systems employ a variety of techniques. Text-based retrieval, for instance, matches user queries with metadata, transcripts, or deep-learning-generated embeddings (e.g., CLIP \cite{radford2021learning}, W2VV++ \cite{li2019w2vv++}), thereby facilitating semantic search, though often at the expense of interpretability. Concept-based methods \cite{chen2020fine, 8100173, fang2022concept, 9334429, 8954166, 10.1007/978-3-030-67835-7_49} use object detection and scene classification models to annotate frames with high-level semantics, yet they are limited by predefined categories and annotation inconsistencies. Similarly, content-based visual search retrieves \cite{amato2021visionevideosearchsystem, gordo2017end, vitrivr} visually similar frames using CNN-based descriptors \cite{tolias2016particularobjectretrievalintegral} but may return semantically irrelevant results without a suitable reference. Even sketch-based and spatial queries, which offer a visual means to express search intent, are challenged by high interpretation variability.

Despite recent advances, current interactive video retrieval systems still exhibit key limitations. First, relying on a single model restricts their ability to capture both broad semantics and fine-grained details \cite{challenge_interactive}. Second, indexing every frame creates redundancy, leading to storage and search inefficiencies. Third, unstable temporal search methods make it difficult to accurately locate event sequences. Finally, conventional reranking overlooks temporal context, resulting in inconsistent rankings. To overcome these challenges, we propose a comprehensive framework featuring four innovations:

\begin{enumerate}
    \item \textbf{Ensemble Search:} By combining coarse-grained models that capture broad semantics with fine-grained models that focus on intricate details, our ensemble approach yields more robust retrieval outcomes.
    \item \textbf{Storage Optimization:} We reduce redundancy by selecting representative keyframes through intelligent deduplication, significantly cutting storage needs without sacrificing search quality.
    \item \textbf{Temporal Search:} Our dual-query mechanism-capturing both start and end points-accurately localizes video segments in chronological order, ensuring stable and interpretable results.
    \item \textbf{Temporal Reranking:} Leveraging contextual information from neighboring frames, our reranking strategy refines candidate orderings to maintain structural coherence and relevance.
    
\end{enumerate}
These contributions address critical shortcomings in existing systems, leading to more precise, efficient, and user-friendly interactive video retrieval. The following sections detail each component, present experimental results, and discuss future implications.

\section{Related Work}
\label{sec:related}

Recent research in video retrieval has evolved along two complementary lines: localized moment retrieval within single videos and corpus-level retrieval that jointly identifies the relevant video and the corresponding moment. In this section, we review key methods spanning both Single Video Moment Retrieval (SVMR) and Video Corpus Moment Retrieval (VCMR), as well as interactive retrieval systems that incorporate human feedback.

\subsection{Video Moment Retrieval}
Video Moment Retrieval aims at localizing a target segment in an untrimmed video based on a natural language query. Early work in this area typically followed a proposal-based paradigm, generating candidate segments which are then ranked based on their relevance to the query \cite{gao2017tall, sun2023video, anne2017localizing, liu2018attentive, chen2018temporally, chen2019semantic}. In contrast, proposal-free approaches directly regress the temporal boundaries by leveraging iterative attention between video frames and query words, often benefiting from transformer architectures \cite{qi2024collaborative, yang2021deconfounded, yuan2019find, lu2019debug, chen2020learning, rodriguez2020proposal, zhang2020temporal, zeng2020dense, lei2021detecting, moon2023query, sun2024tr, gordeev2024saliency}. 

The scope of video retrieval has further expanded to Video Corpus Moment Retrieval (VCMR), where the task is to first identify candidate videos from a large collection and subsequently localize the relevant moment within the selected video. Methods for VCMR are generally divided into one-stage and two-stage approaches \cite{yoon2022selective, zhang2020hierarchical, lei2020tvr, li2020hero, zhang2021video, hou2021conquer, zhang2023video}. One-stage methods jointly perform retrieval and localization in an end-to-end manner; for instance, HERO \cite{li2020hero} employs a hierarchical transformer-based encoder that integrates visual and textual cues across modalities. Two-stage methods, on the other hand, first retrieve a set of candidate videos based on global text–video similarity and then apply fine-grained localization on each candidate. CONQUER \cite{hou2021conquer} exemplifies this approach by introducing a query-aware ranking mechanism that benefits from enhanced interactions between the query and video content, thereby offering improved scalability for large-scale video retrieval.

Moreover, the emergence of large language models has influenced both SVMR and VCMR. Recent studies integrate video understanding and moment retrieval into a next-token prediction framework \cite{huang2024vtimellm, wang2024hawkeye, guo2024vtg, wang2024grounded}. Additionally, generative techniques have been explored; for instance, MomentDiff \cite{li2023momentdiff} formulates moment retrieval as a diffusion process that iteratively refines random temporal proposals into the correct segment.

\subsection{Interactive Retrieval Systems}
While fully automated retrieval pipelines have made significant progress, they often struggle with complex queries and long video inputs. To address these limitations, interactive retrieval systems incorporate human-in-the-loop strategies. Benchmarks such as the Video Browser Showdown (VBS) have spurred research on multimodal interactive methods that combine text, sketches, filters, and example frames. For example, \cite{ma2022interactive} proposed a reinforcement learning-based framework that learns from user feedback to navigate large video corpora, while \cite{liang2023simple, nguyen2024improvinggeneralizationvisualreasoning} introduced a question-answering-based interactive system that simulates user interactions using a VideoQA model. These interactive approaches demonstrate that iterative refinement can substantially improve retrieval performance, making them particularly promising for real-world applications.

\section{Method}
\label{sec:method}
\begin{figure*}
    \centering
    \includegraphics[width=0.75\linewidth]{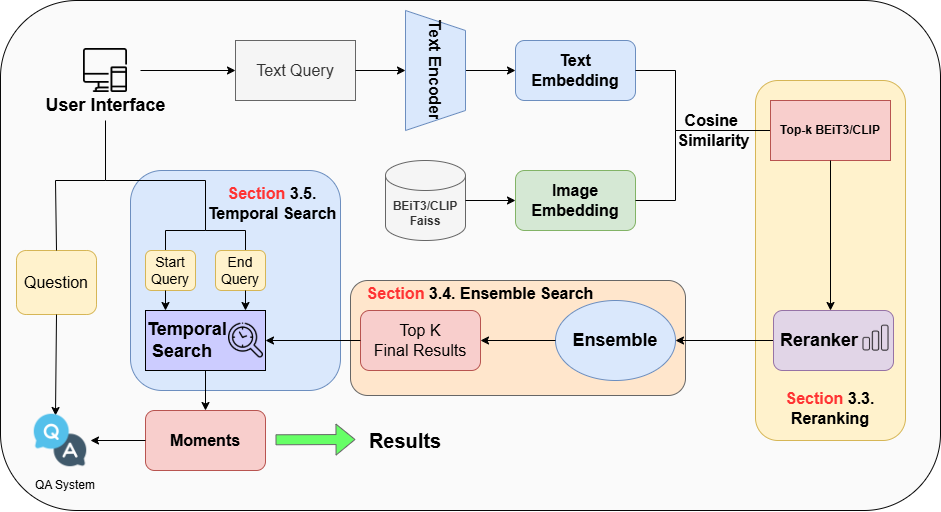}
    \caption{Overview of our iteractive retrieval system. The system ranks the top-k results through a reranking module (\textbf{Section}~\ref{subsec:rerank} before passing them to an ensemble module (\textbf{Section}~\ref{subsec:ensemble}) for final selection. The temporal search module (\textbf{Section}~\ref{subsec:temporal}) refines the results by identifying the most relevant time segments, ensuring the retrieval aligns with the query's temporal context. The final output consists of the most relevant moments, providing accurate answers based on the keyframe range.}
    \label{fig:enter-label}
\end{figure*}

\subsection{Problem Definition}

Holistic video understanding remains challenging, as most existing methods focus on short video segments despite the growing prevalence of long-form content spanning minutes to hours. Given a corpus of untrimmed videos $\mathcal{V}$ and a textual query $q$, the objective of Video Corpus Moment Retrieval (VCMR) is to identify the specific moment $m^* = (t^s, t^e)$ that best aligns with $q$, where $t^s$ and $t^e$ denote the start and end timestamps, respectively:
\begin{equation}
    m^* = \underset{m}{\arg\max} \; P(m \mid q, \mathcal{V}).
\end{equation}

The VCMR process comprises two stages: (i) retrieving candidate moments $m$ from videos in the corpus, and (ii) accurately localizing the optimal moment $m^*$ within the selected video $v^*$. To further enhance the system's capabilities, we incorporate a Question Answering (QA) component, allowing users to interact with localized moments. This facilitates refined responses by leveraging additional context from the targeted segment.

Traditional video-level retrieval methods often struggle to capture the fine-grained temporal details necessary for accurate localization. To overcome this, we shift to an image-level retrieval approach, treating individual frames as the basic retrieval units. Although this method lacks immediate temporal context, we mitigate this shortcoming using a combination of reranking (\textbf{Section}~\ref{subsec:rerank}), ensemble search (\textbf{Section}~\ref{subsec:ensemble}), and temporal modeling (\textbf{Section}~\ref{subsec:temporal}). These components collectively ensure that the retrieved frames maintain both semantic relevance and temporal coherence.


\begin{figure}
    \centering
    \includegraphics[width=0.9\linewidth]{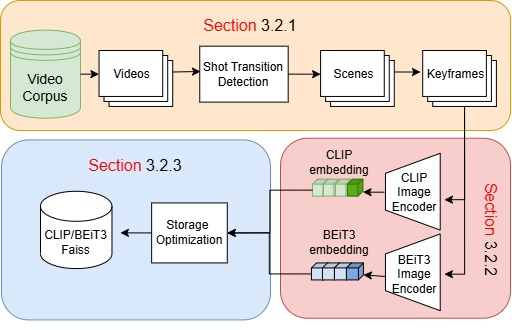}
    \caption{Videos are segmented, deduplicated with cosine embeddings, and stored in the FAISS index.}
    \label{fig:enter-label}
\end{figure}

\subsection{Data storage}
For efficient image-level video retrieval, an optimized data storage strategy is critical for efficient retrieval performance. 
By selecting a minimal but representative set of frames, we reduce redundancy, making retrieval more efficient without compromising accuracy. In this section, our data storage comprising three stages: keyframe selection \textbf{\ref{subsubsec:keyframe_selec}}, feature extraction \textbf{\ref{subsubsec:feat}}, and storage optimization \textbf{\ref{subsubsec:storage}}.

\subsubsection{Keyframe Selection} \label{subsubsec:keyframe_selec}
To extract keyframes from videos, we utilize TransNetV2\cite{soucek2024transnet}, a fast and accurate deep learning model designed for scene transition detection. TransNetV2\cite{soucek2024transnet} is a CNN and RNN hybrid that excels in detecting hard cuts and gradual transitions within video sequences. It processes input frames sequentially and identifies transition probabilities, enabling precise segmentation of scenes.

Once scene transitions are detected, we select keyframes for each scene. Instead of storing all frames, we sample four evenly spaced frames within each detected scene based on their frame indices. This strategy ensures a diverse yet compact representation of the scene, reducing storage requirements while preserving essential visual information.

\subsubsection{Feature Extraction} \label{subsubsec:feat}
To enhance retrieval accuracy, we employ powerful feature extractors, specifically BEIT3\cite{wang2023image} and CLIP\cite{radford2021learning}. These models have demonstrated superior performance across multiple benchmark datasets, making them well-suited for video retrieval tasks. These models are particularly well-suited for our task due to their ability to generate robust, high-dimensional feature embeddings, capturing intricate visual relationships that traditional CNN-based models might overlook. 


\subsubsection{Storage Optimization} \label{subsubsec:storage}
Despite selecting only four keyframes per scene, redundant frames may still exist, leading to suboptimal memory usage and slower retrieval times. We implement a duplicate removal algorithm based on feature similarity to address this.

We compute the cosine similarity between keyframes within each scene using the extracted feature embeddings from BEIT3 and CLIP. If a frame has a cosine similarity score greater than 0.9 with any other frame in the same scene, it is considered a near-duplicate and is removed. This process significantly reduces storage redundancy while maintaining retrieval efficiency. The \textbf{Algorithm} \ref{alg:frame_filter} describes all of the above.

\begin{figure}[!h]
  \includegraphics[width=0.5\textwidth]{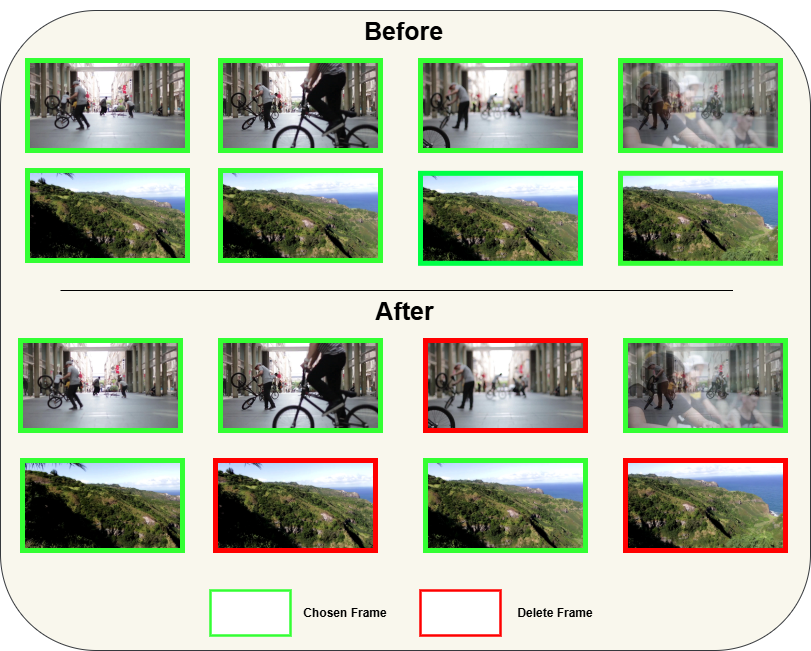}
  \caption{Before and after frame filtering.}
  \label{fig:frame_filter}
\end{figure}

By integrating efficient keyframe selection, feature extraction, and storage optimization, our approach ensures a highly scalable and accurate image-level video retrieval system. The effectiveness of our storage optimization strategy is visually demonstrated in the \textbf{Figure \ref{fig:frame_filter}}, highlighting the impact of duplicate removal on the dataset efficiently. 


%

\begin{algorithm}[!h]
\caption{Frame Filtering}
\label{alg:frame_filter}
\begin{algorithmic}[1]
\For{each \textbf{group}, \textbf{values} in \texttt{grouped\_videos}}
    \State Load scene boundaries from file
    \State Initialize \texttt{id} = 0 and \texttt{frame\_in\_scene} = []
    \State Set $S \gets 0.9$
    \For{each \textbf{key}, \textbf{path} in \textbf{values}}
        \If{current frame is within the scene boundary}
            \State Add frame to \texttt{frame\_in\_scene}
        \Else
            \State Compute embeddings: $f$ frame embedding
            \State Compute similarity: $E = \texttt{sim}(f_1, f_2)$
            \State Identify redundant frames (if $E > S$)
            \State Remove redundant frames
            \State Move to the next scene (\texttt{id} += 1)
            \State Reset \texttt{frame\_in\_scene}
        \EndIf
    \EndFor
\EndFor
\end{algorithmic}
\end{algorithm}

\subsection{Reranking} \label{subsec:rerank}
As described in \textbf{Algorithm~\ref{alg:neighborscore}}, the neighbor score aggregation method should be seen as a strong keyframe selection approach due to the enhancement of both stability and temporal relevance. Since a shot detection model has previously been employed to retrieve the keyframes, we believe that the areas surrounding a keyframe are most likely to possess visual and semantic features in common or correspond to temporal shifts within a shot. Hence, this approach guarantees reliable keyframe selection when local neighbors show stable visual similarity and ensures temporal coherence when the queried content depicts motion that covers some of the adjacent frames.

The trade-off of stability and temporal relevance is depicted in the \textbf{Algorithm~\ref{alg:neighborscore}}. The aggregated score reinforces the candidate keyframe on stable neighbors - both semantically and scores-wise responsible for a strong representation of the scene. Aggregated against the neighbors based on a temporally descriptive query, it explicitly expresses the context of any dynamic for which a query illustrates. Thus, extreme situations like sudden changes in the scene would be handled by the conditional score check that would filter out the irrelevant or missing part of the contribution and thus the method could achieve some decent robustness. 
{\setstretch{0.9} 
\begin{algorithm}
\caption{Neighbor Score Aggregation}
\label{alg:neighborscore}
\begin{algorithmic}[1]
\Require Indcies I, Query Q, 
    \Function{AggregateNeighborScores}{\texttt{I}, \texttt{Q}}
        \State Initialize dictionary \texttt{aggregated\_score} A
        
        \For{each $idx$ in \texttt{I}}
            \State $key \gets$ Convert $idx$ to integer
            \State $\texttt{neighbors} \gets$ \Call{GetNeighbors}{$key$}
            \State $total\_score \gets 0$
            
            \For{each $neighbor$ N in \texttt{neighbors}}
                \State \texttt{score} $\gets$ \Call{ComputeScore}{\texttt{N}, \texttt{Q}}
                \If{$score \neq$ None}
                    \State Append $neighbor$ to \texttt{indices}
                    \State $total\_score \gets total\_score + score$
                \EndIf
            \EndFor
            
            \State \Call{UpdateScores}{\texttt{A}, $key$, $total\_score$}
        \EndFor
        
        \State \texttt{sorted\_scores} $\gets$ \Call{Sort}{\texttt{A}, \texttt{descending}}
        \State \Return \texttt{sorted\_scores}
    \EndFunction
\end{algorithmic}
\end{algorithm}
} 

\subsection{Ensemble Search} \label{subsec:ensemble}
Image-text retrieval requires both fine-grained visual understanding and coarse-grained conceptual alignment. BEiT-3 \cite{wang2023image} excels in detailed image-text alignment but may overlook broader context, while CLIP \cite{radford2021learning} performs well in zero-shot retrieval but struggles with fine-grained distinctions. These differences stem from their training: BEiT-3 focuses on comprehensive semantic understanding, whereas CLIP learns condensed representations from limited text. To balance precision and recall, we propose an ensemble search method that combines their strengths, improving retrieval robustness across diverse queries.

\begin{algorithm}[!h]
\caption{Ensemble Search}
\label{alg:ensemble_search}
\begin{algorithmic}[1]
\Require query (string), model\_configs (list of (model\_name, weight, use\_flag))
\Ensure ranked\_results (list of (index, score))

\State Initialize $\text{score\_dict}$ as empty dictionary
\State Normalize weights: $\sum w_i = 1$ for selected models

\ForAll {$(\text{model\_name}, w, \text{use\_flag}) \in \text{model\_configs}$}
    \If {use\_flag}
        \State Load model and processor for \text{model\_name}
        \State $e \gets \text{EncodeText}(\text{model\_name}, \text{query})$
        \State $I, S \gets \text{Search}(\text{model\_name\_idx}, e, M=50)$
        \State $S_{\max} \gets \max(S)$
        \ForAll {$(i, s) \in (I, S)$}
            \State $\text{score\_dict}[i] \mathrel{+}= \frac{s}{S_{\max}} \times w$
        \EndFor
    \EndIf
\EndFor

\State $\text{ranked\_results} \gets \text{Sort}(\text{score\_dict}, \text{descending})$
\State \Return $\text{ranked\_results}$

\end{algorithmic}
\end{algorithm}






Our ensemble search methodology is formally outlined in \textbf{Algorithm~\ref{alg:ensemble_search}}. The process begins by encoding the query text with both models using their tokenization methods and normalizing the results to unit length to produce text embeddings. These model-specific query embeddings are then used to retrieve the top M results from each corresponding index. To effectively combine these results, we normalize the similarity scores from each model by their respective maximum values, thereby preventing disparities in scale. By applying weighting factors during ensemble search, the approach reflects the relative contribution of each model. The weighted scores are aggregated by each image identifier, with scores being summed when an image appears in both result sets. The final step incorporates sorting these aggregated scores in descending order to produce a comprehensive list that both precisely detailed and broadly relevant to the query, overcoming the limitations of each model in isolation.

\subsection{Temporal Search} \label{subsec:temporal}


Temporal video search is challenging due to the dynamic nature of visual content and frame dependencies. Unlike static image retrieval, it must identify not just relevant frames but also their temporal range for accurate event localization. A single frame often gives only a rough estimate without clear temporal boundaries.
To address this issue, we propose a bidirectional temporal search strategy that refines retrieval precision by leveraging additional queries, depicted in \textbf{Algorithm ~\ref{alg:frame_pair_selection}}. We assume that the initially retrieved and reranked input frame corresponds to the correct reference frame, and our objective is to localize the segment that best aligns with the given query. To achieve this, we conduct a bidirectional search, extending to the left of the input frame index until either 20 relevant frames are identified or the similarity score falls below an acceptable threshold. The same approach is applied symmetrically to the right. Following this, we determine the optimal frame pair by selecting two frames that exhibit the highest similarity scores with the respective queries while ensuring that their temporal distance, including the input frame, does not exceed a predefined constraint, denoted as $gap_C$. This approach effectively refines temporal retrieval by incorporating local context while maintaining alignment with the query semantics.

\begin{algorithm}[!h]
\caption{Temporal Frame Pair Selection}
\label{alg:frame_pair_selection}
\begin{algorithmic}[1]
    \Function{FindBestFramePair}{query\_1, query\_2, input\_frame, index, img\_path, gap\_C}
        \State Identify the video and frame ID from \textit{input\_frame}
        \State Initialize a list for relevant \textbf{left} and \textbf{right} frames
        \State Set a similarity \textbf{threshold} to filter frames
        
        \While{Frame is relevant and not exceeding limit}
            \State Compute similarity with \textit{query\_1}
            \State Stop if similarity is too low
            \State Add frame to \textbf{left list} and move left
        \EndWhile
        
        \While{Frame is relevant and not exceeding limit}
            \State Compute similarity with \textit{query\_2}
            \State Stop if similarity is too low
            \State Add frame to \textbf{right list} and move right
        \EndWhile
        
        \State Collect all candidate frames from left, current, and right
        \State Find the best pair of frames that maximizes combined similarity
        \State Ensure the frames satisfy the temporal constraint (gap\_C)
        
        \State \Return the best matching frame pair
    \EndFunction
\end{algorithmic}
\end{algorithm}

\subsection{Interactive Video Retrieval System}
\begin{figure}[t]
    \centering
    %
    \begin{subfigure}[b]{\linewidth}
        \centering
        \fbox{\includegraphics[width=\linewidth]{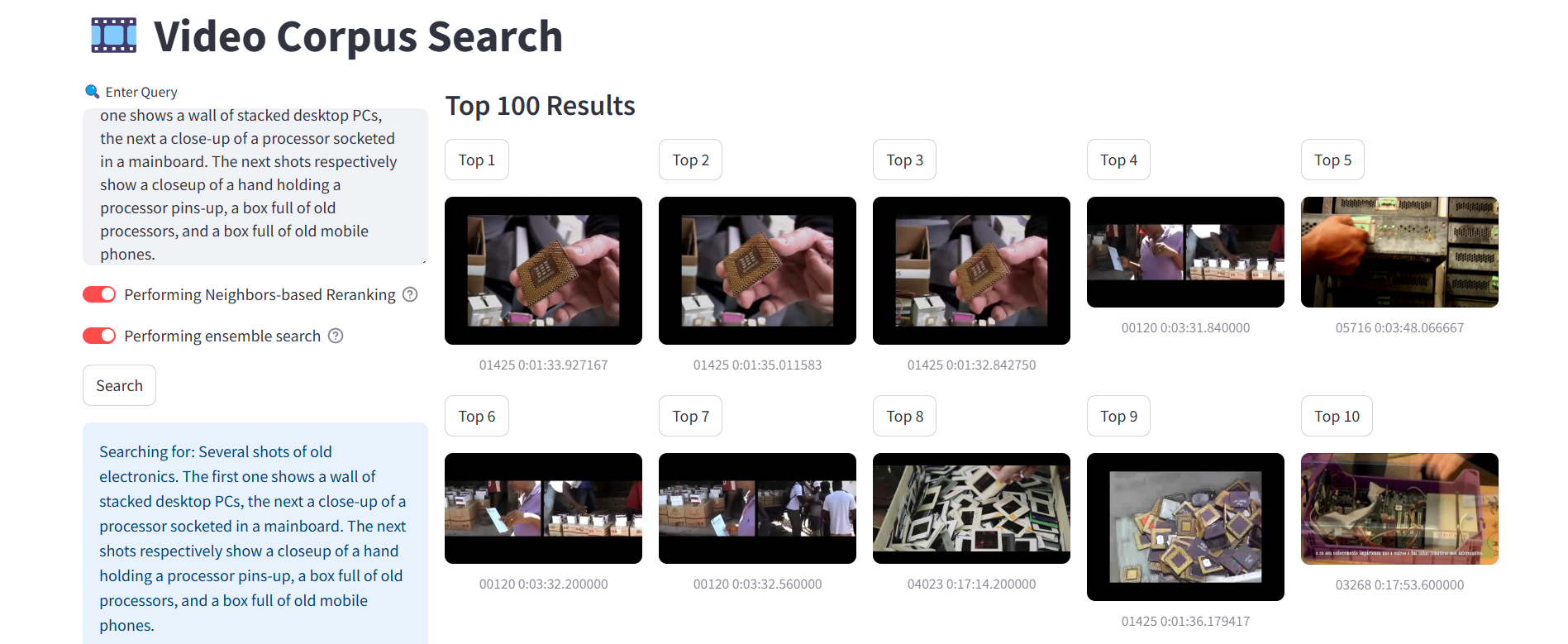}}
        \caption{Overall User Interface}
        \label{fig:search1}
    \end{subfigure}
    \hfill
    %
    \begin{subfigure}[b]{\linewidth}
        \centering
        \fbox{\includegraphics[width=\linewidth]{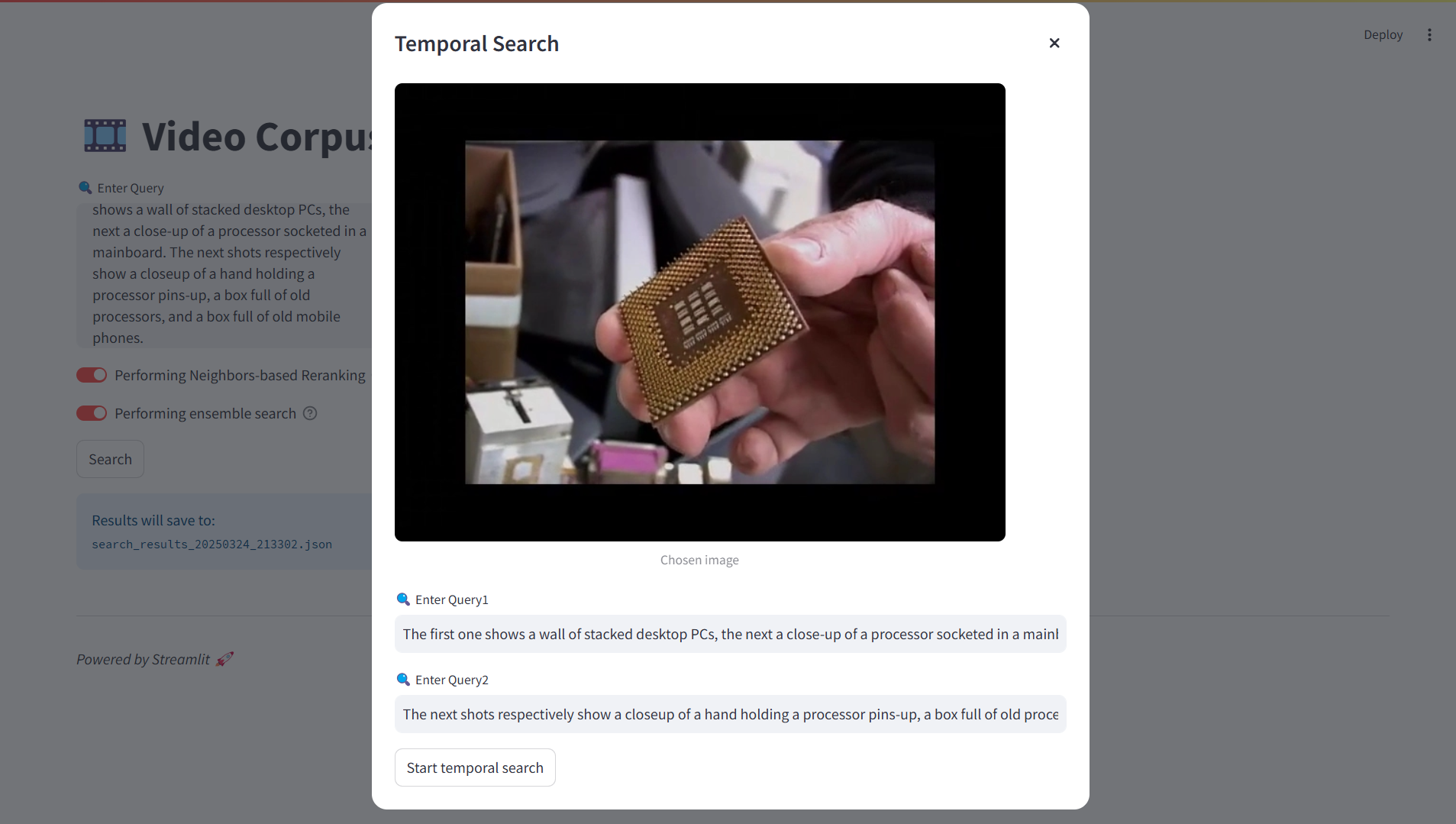}}
        \caption{Temporal Search User Interface}
        \label{fig:search2}
    \end{subfigure}
    \hfill
    %
    \begin{subfigure}[b]{\linewidth}
        \centering
        \fbox{\includegraphics[width=\linewidth]{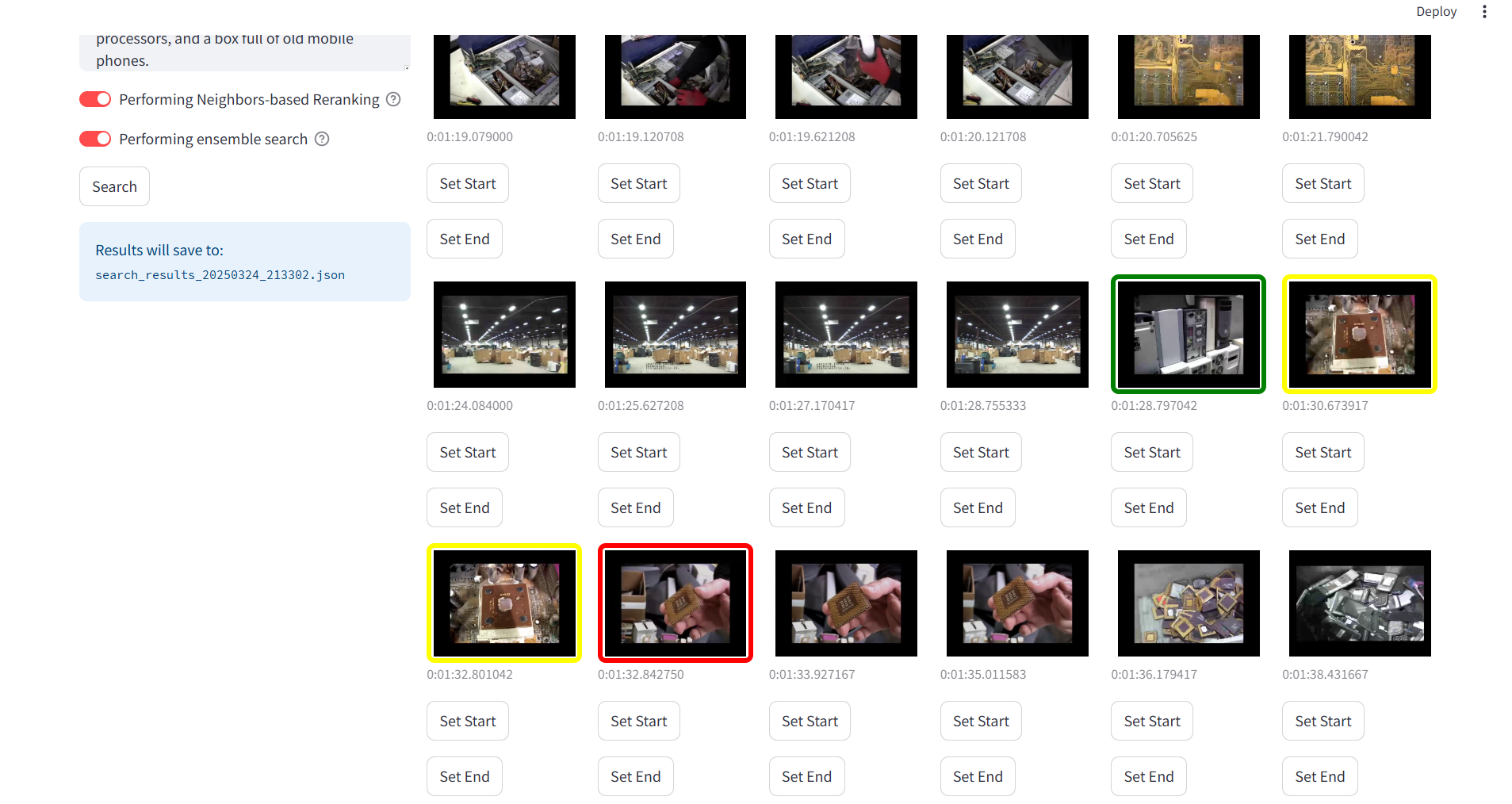}}
        \caption{Boundary Selection}
        \label{fig:search3}
    \end{subfigure}
    \caption{The UI for Interactive Retrieval System. The selected start frame will be annotated as a frame with green border, whereas the end frame will be annotated as a frame with red border}
    \label{fig:multi_sub}
\end{figure}

In this section, we introduce our \emph{Interactive Video Retrieval System}, designed to handle two core tasks: \emph{Moment Retrieval} and \emph{Video Question Answering (QA)}. The system features a user-friendly interface where queries can be entered, search strategies can be selected, and results can be iteratively refined. \textbf{Figures~\ref{fig:search1}, \ref{fig:search2}, and \ref{fig:search3}} illustrate various stages of user interaction within the system.

\subsubsection{Moment Retrieval}
\label{sec:ivs_mr}

As shown in \textbf{Figure~\ref{fig:search1}}, the user can enter a free-form text query in the designated text box. Two retrieval strategies, \emph{neighbors-based reranking} and \emph{ensemble search}, are provided to adjust how the system prioritizes candidate frames. Neighbors-based reranking refines the initial results by leveraging local similarities between frames, while ensemble search combines multiple search strategies to enhance retrieval robustness.

\textbf{Top-100 Keyframe Display.}
After the user submits the query, the system returns up to 100 keyframes that best match the textual description (\textbf{Figure~\ref{fig:search1}}). Each keyframe is displayed with a timestamp, allowing users to quickly assess and compare different candidates.

\textbf{Temporal Search with Dual Queries.}  
To accurately pinpoint the desired video segment, users provide \textbf{two separate textual descriptions} (\textbf{Figure \ref{fig:search2}}): one describing the \textbf{start} of the moment and another describing the \textbf{end}. Based on these two mini-queries, the system suggests a start frame and an end frame within the relevant video. \textbf{Figure~\ref{fig:search3}} shows an example interface where the system highlights the proposed start frame in green and the proposed end frame in red. Users can review these suggestions and adjust them if necessary to refine the moment boundaries.

\textbf{Finalizing the Moment.}  
Once satisfied with the suggested or adjusted frames, the user confirms the selection. The system then extracts and logs the identified segment, which can subsequently be used for more detailed examination or for tasks such as QA.

\subsubsection{Video Question and Answering (QA)}
\label{sec:ivs_qa}

After the system identifies the relevant segment via \emph{temporal search}, users can perform \textbf{Video Question Answering (QA)} by observing the extracted frames. Unlike automated QA systems, our approach relies on users to examine the displayed segment and derive the most suitable answer based on their own observations.


Observation-based QA has a number of benefits, such as increased accuracy through the use of direct observation instead of assumptions, increased reliability through real-time data gathering, and greater flexibility in adapting to changing environments. It also helps detect subtle anomalies that VideoQA models might miss, leading to more effective decision making and better overall quality assurance.

\section{Experimental Results}
\label{sec:experiment}

\subsection{Known-Item Search}
\begin{figure}[t]
    \centering
    %
    \begin{subfigure}[b]{\linewidth}
        \centering
        \fbox{\includegraphics[width=\linewidth]{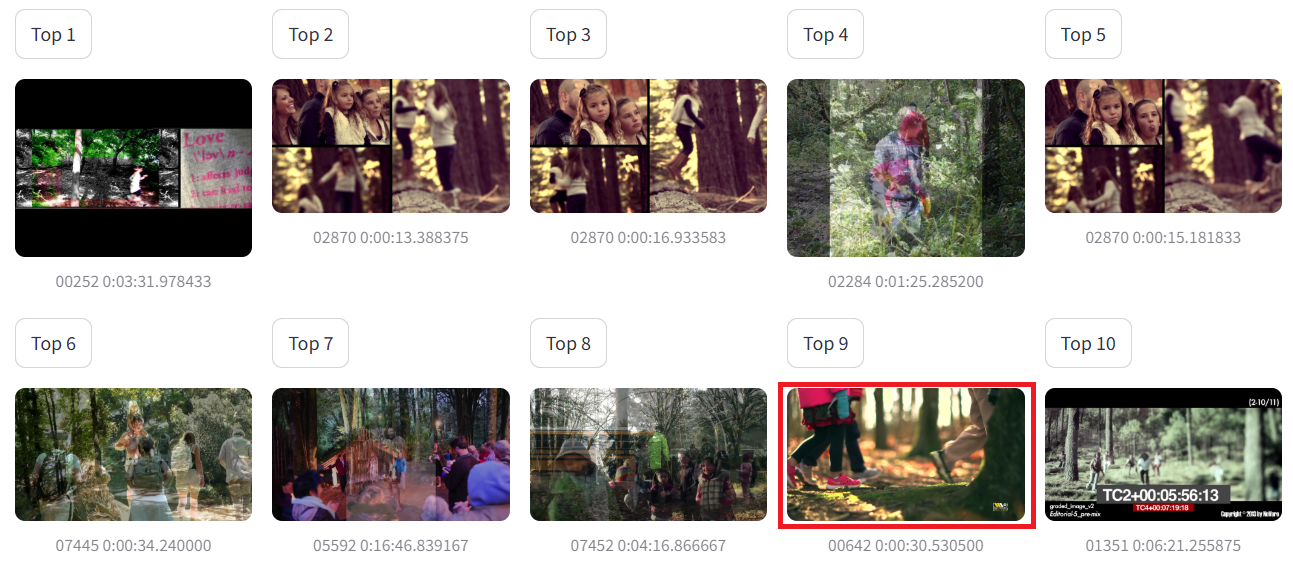}}
        \caption{Using only BeIT-3 features (no rerank or ensemble).}
        \label{fig:none}
    \end{subfigure}
    \hfill
    %
    \begin{subfigure}[b]{\linewidth}
        \centering
        \fbox{\includegraphics[width=\linewidth]{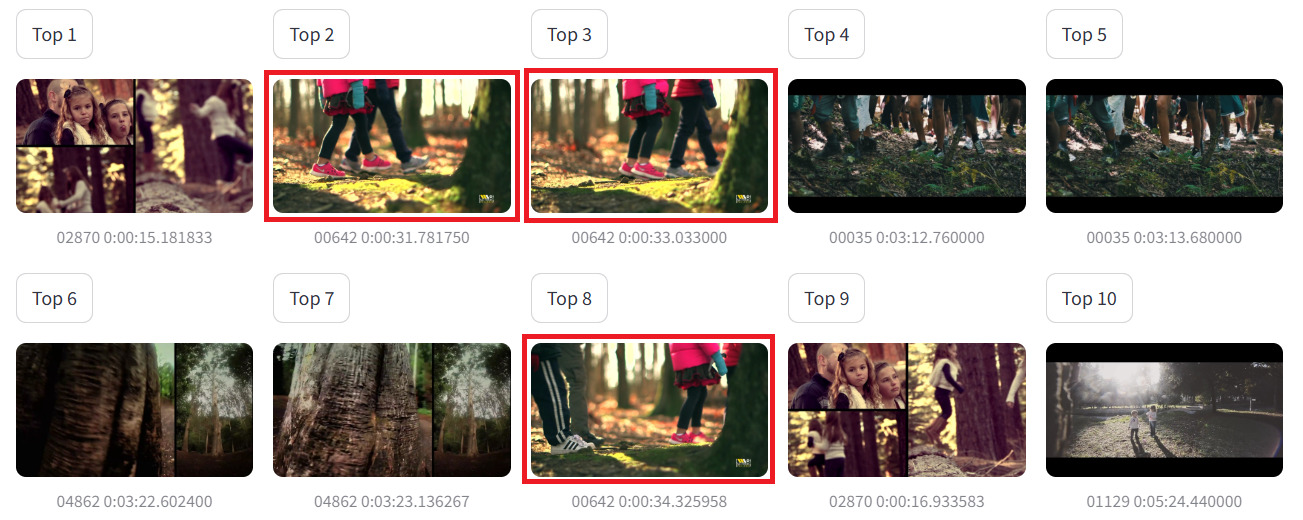}}
        \caption{Neighbors-based reranking approach.}
        \label{fig:rerank}
    \end{subfigure}
    \hfill
    %
    \begin{subfigure}[b]{\linewidth}
        \centering
        \fbox{\includegraphics[width=\linewidth]{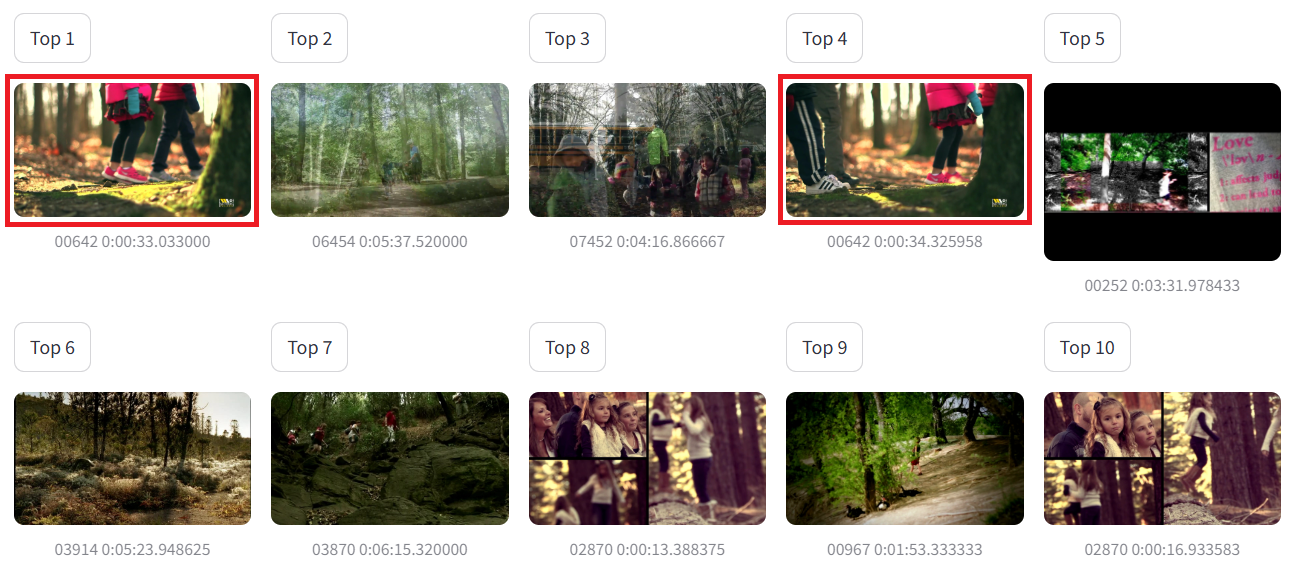}}
        \caption{Ensemble search (BeIT-3 + OpenCLIP).}
        \label{fig:ensemble}
    \end{subfigure}
    %
    \begin{subfigure}[b]{\linewidth}
        \centering
        \fbox{\includegraphics[width=\linewidth]{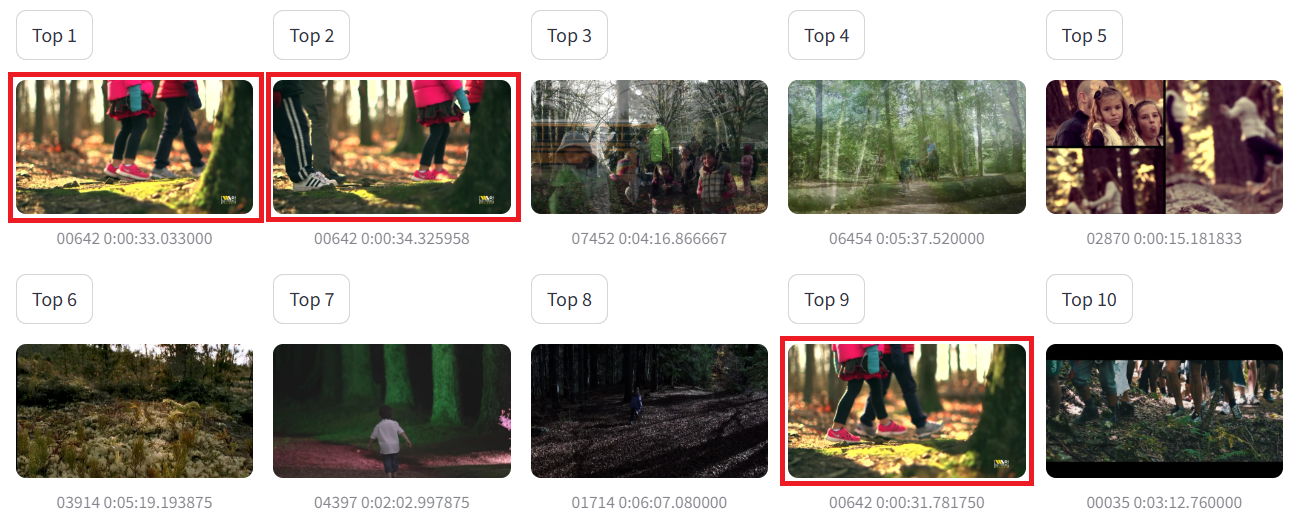}}
        \caption{Combining both ensemble and reranking strategies.}
        \label{fig:ensemble+rerank}
    \end{subfigure}
    \caption{Experimental results for Known-Item Search. The actual target frame is highlighted with a red box.}
    \label{fig:multi_sub}
\end{figure}

\subsubsection{Reranking and Ensemble Search}

We evaluate our system on a \emph{Known-Item Search} task, aiming to retrieve specific frames from a video based on a user-defined query. Three retrieval strategies are examined:

\begin{enumerate}
    \item \textbf{Single-Model (BeIT-3):} Relying solely on the BeIT-3 feature extractor to match frames against the query.
    \item \textbf{Neighbors-Based Reranking:} Refining the initial candidate list by leveraging stable local neighborhoods, thereby boosting frames that are contextually consistent.
    \item \textbf{Ensemble (BeIT-3 + OpenCLIP):} Combining two feature extractors to incorporate multiple “views” of the query for greater robustness.
    \item \textbf{Combining Ensemble and Reranking:} Integrating both ensemble search and reranking to maximize robustness, handling variations in difference perspectives.
\end{enumerate}

\noindent
\textbf{Query Example:}
\emph{``Two scenes in a forest. In the first shot, several people are walking as sunlight shines on the ground, and only the lower body of the people is visible. On the right side of the frame there is a tree covered in green moss. In the second shot, we know that they are children walking through the woods.''}

\textbf{Single-Model Baseline (Figure~\ref{fig:none}).}
Using only BeIT-3 features provides a baseline for performance. While the system identifies one correct frame, it is ranked relatively low, implying that a single-model approach struggles with nuanced scene details-such as partially visible bodies or the presence of a moss-covered tree. This limitation underscores the need for strategies that capture both fine-grained and coarse-grained cues.

\textbf{Neighbors-Based Reranking (Figure~\ref{fig:rerank}).}
To address the shortcomings of the baseline, we introduce a neighbor-based reranking step. By examining local neighborhoods in feature space, the system promotes frames that share stable visual cues, thus better aligning with the query context. The results show a notable increase in top-ranked correct frames, indicating that spatial/temporal consistency plays a crucial role in distinguishing truly relevant frames from visually similar distractors. 

\textbf{Ensemble Search (Figure~\ref{fig:ensemble}).}
Next, we integrate BeIT-3 and OpenCLIP into an ensemble to combine multiple “views” of the data. BeIT-3 excels at detailed, fine-grained matching, while OpenCLIP offers strong semantic alignment. Merging these representations often propels correct frames into top-1 or top-5 positions, enhancing the system’s ability to detect subtle scene elements and high-level thematic cues simultaneously. In practice, this synergy is particularly beneficial for queries that describe both specific objects (e.g., mossy tree) and overarching context (children walking).

\textbf{Combining Ensemble and Reranking (Figure~\ref{fig:ensemble+rerank}).}
Finally, uniting both ensemble search and neighbors-based reranking yields the most robust results. By first leveraging diverse feature extractors and refining through local coherence, the system effectively handles challenges such as partial occlusions, lighting shifts, and complex motion. Across all trials, frames marked with red bounding boxes consistently surface at the top ranks, demonstrating the combined advantage of ensemble and reranking over the single-model baseline in terms of precision and stability.

\subsubsection{Temporal Search}
\begin{figure}[!h]
    \centering
    \fbox{\includegraphics[width=\linewidth]{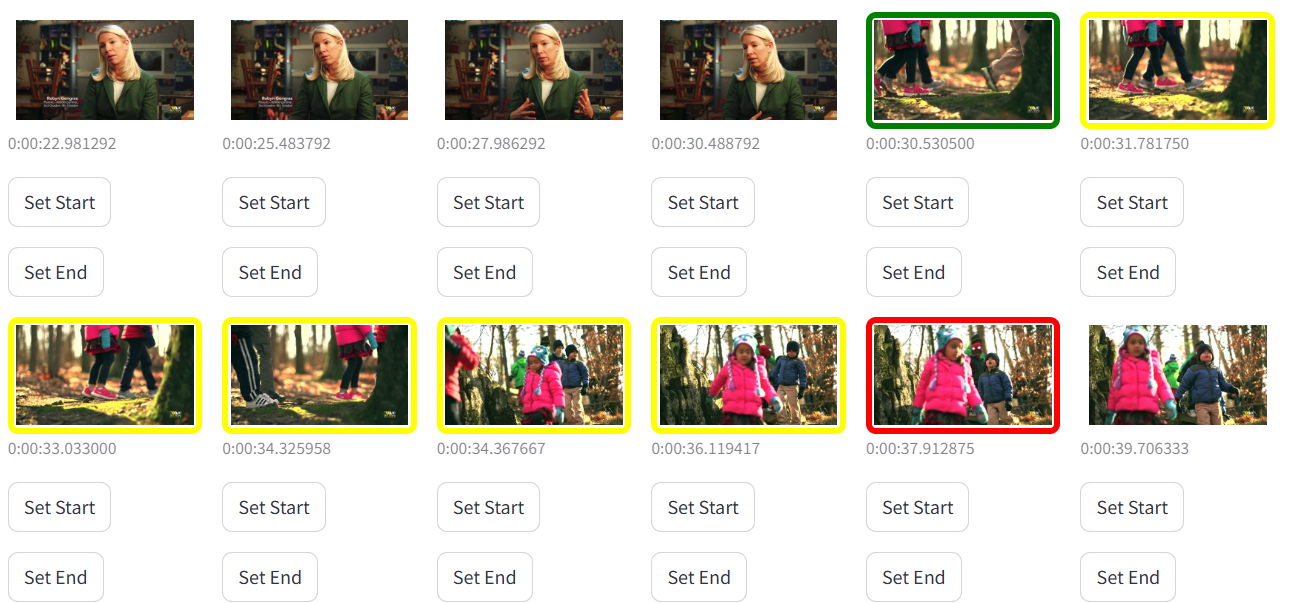}}
    \caption{Temporal Search Example. The start and end frames are enclosed by a green and red box, respectively.}
    \label{fig:temporal}
\end{figure}

Building on the enhanced list of candidate frames from the image-based retrieval stage, we introduce a \textbf{temporal search} mechanism. Unlike single-frame retrieval, temporal search aims to locate a continuous video segment that evolves from a “start” description to an “end” description, thus capturing the natural progression of events. Specifically, we use two distinct textual queries:

\begin{itemize}
    \item \textbf{Start Query:} ``In the first shot, several people are walking, and on the right side of the frame there is a tree covered in green moss; the camera only shows their lower bodies.''
    \item \textbf{End Query:} ``In the second shot, we learn that they are children walking through the woods.''
\end{itemize}

By incorporating temporal constraints-such as the chronological order and semantic linkage between two events-our system more accurately reconstructs the full narrative of the user’s query. This approach is grounded in the notion that many video concepts (e.g., characters entering or leaving the frame, environment changes) are inherently sequential and cannot be fully represented by a single static image. Consequently, temporal search delivers a more holistic view, allowing users to observe how a scene unfolds between the designated start and end points.

As demonstrated in \textbf{Figure~\ref{fig:temporal}}, this dual-query approach consistently yields a focused segment containing the relevant frames between the start and end descriptions. Even in videos featuring gradual transitions or subtle camera movements, the temporal search mechanism effectively localizes the moment of interest. In practice, users reported that specifying two queries not only helped them retrieve more accurate results but also made the retrieval process feel more natural, as it closely mirrored how people describe events in everyday conversation (i.e., “It starts when X happens and ends when Y occurs.”).

\subsection{Question-Answering}
\begin{figure}[!h]
    \centering
    \fbox{\includegraphics[width=\linewidth]{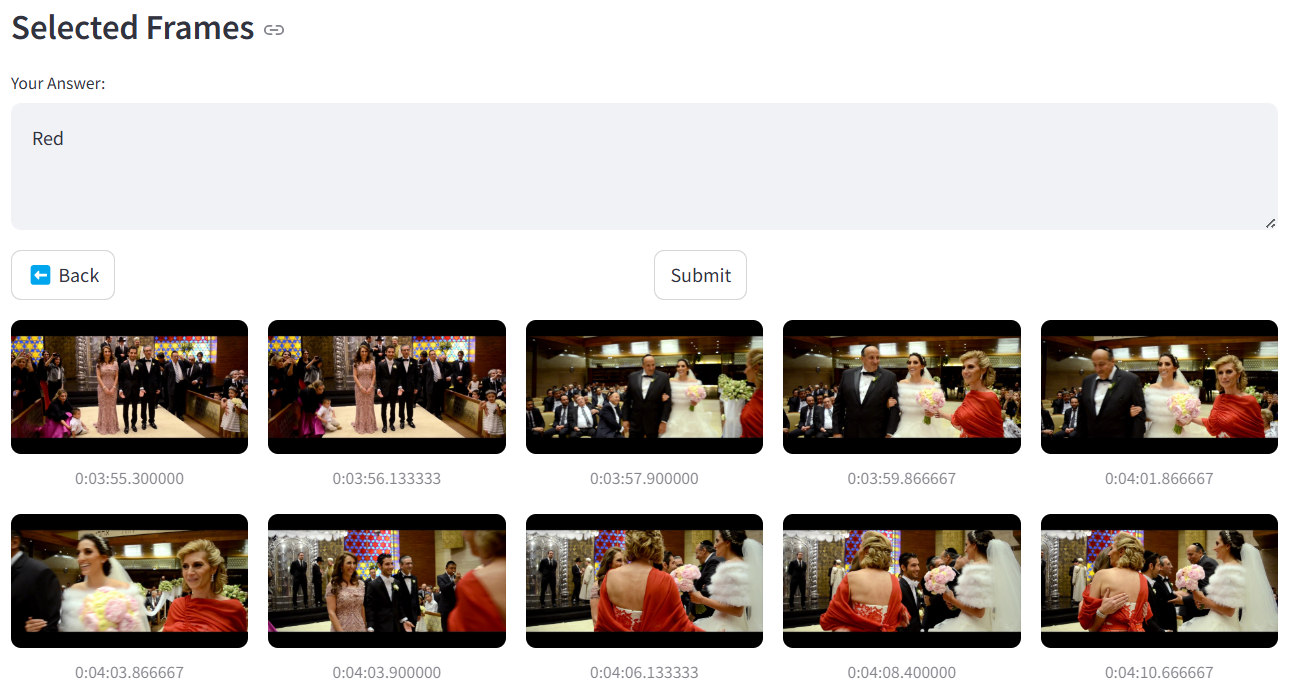}}
    \caption{Example QA interface. After retrieving the relevant moment, the system displays a sequence of frames that help the user answer the posed question.}
    \label{fig:qa}
\end{figure}

In the \textbf{Question-Answering (QA)} task, the user first employs our Known-Item Search approach to pinpoint the specific moment of interest in a long video. Once that segment is located, the system presents a series of sequential frames spanning from the user-defined start to the end of the event. This detailed, frame-by-frame visualization allows the user to observe contextual and visual cues that would otherwise be lost in single-frame retrieval.

\noindent
\textbf{Query Example:}
\emph{``The groom, flanked by family, awaits his bride with excitement and anticipation. The radiant bride walks down the aisle, escorted by her proud parents, carrying a bouquet of fresh flowers. The bride approaches the groom, as guests witness this special moment in a beautifully adorned ceremony hall. What color is the bride's mother's dress?''}

Once the user has inspected the relevant frames as shown in \textbf{Figure \ref{fig:qa}}, they input their final answer (e.g., \emph{“Red”} for the mother’s dress color) into the answer box. The system then records the user’s response along with any selected frames. This mechanism not only boosts transparency but also aids future analysis, allowing users or evaluators to verify how a particular answer was derived.

\section{Conclusion}
\label{sec:conclusion}


In this work, we propose a unified framework for interactive video retrieval that addresses the challenges of long-form content. By integrating ensemble search, storage optimization, temporal search, and temporal reranking, our approach overcomes the limitations of existing systems, enhancing both accuracy and efficiency. Through a combination of coarse- and fine-grained retrieval models, our method ensures precise content identification while minimizing redundancy. Our framework demonstrates the promise of human-computer collaboration, offering a scalable solution for content-based video search and multimedia analysis, with strong performance on known-item and question-answering tasks.
{
    \small
    \bibliographystyle{ieeenat_fullname}
    \bibliography{main}
}


\end{document}